\documentclass[11pt]{article}

\usepackage{acl}

\usepackage{times}
\usepackage{latexsym}

\usepackage[T1]{fontenc}

\usepackage[utf8]{inputenc}
\usepackage{amsmath}
\usepackage{todonotes}
\usepackage[most]{tcolorbox}
\usepackage{booktabs, multirow} 
\usepackage{soul}
\usepackage{xcolor,colortbl} 
\usepackage{changepage,threeparttable} 
\usepackage{microtype}
\usepackage{makecell}
\usepackage{inconsolata}
\usepackage{CJKutf8}

\usepackage{graphicx}

\usepackage{newunicodechar}
\newcommand{\lowblock}{\rule[0.3ex]{1em}{0.15ex}}
\newunicodechar{▁}{\lowblock}
\title{Qwenfant: Train Big Character-level BabyLMs on Concise Parallel Text}

\author{Lukas Edman$^{1,2}$\qquad Alexander Fraser$^{1,2,3}$ \vspace{.2cm}\\ 
$^{1}$School of Computation, Information and Technology, TU Munich \\
$^{2}$Munich Center for Machine Learning \\
$^{3}$Munich Data Science Institute \\
\vspace{.1cm} {\tt \small lukas.edman@tum.de}
}

\begin{document}
\maketitle
\begin{abstract}


This work concerns our submission for the \textsc{Multilingual} track of the 2026 BabyLM Challenge.\footnote{\url{https://github.com/Leukas/babylm26}}
We improve on our baseline using data that has been summarized and translated by an LLM, showing that higher-quality data can come from synthetic sources. We further improve with 2 simple changes: using a larger model, and using a character-level vocabulary. Both of these changes require LoRA-style fine-tuning for realizing their full potential. 
We dub our final Qwen3-based model Qwenfant, which, at the time of submission, ranks 1st in the \textsc{Multilingual} track.

\end{abstract}

\section{Introduction}

Language Models (LMs) are generally considered to be much less data-efficient than humans, requiring orders of magnitude more data to achieve similar performance.
However the nature of the data seen by humans is vastly different from that typically seen by these LMs. The majority of LM training data is mined from the web, with little to no regard to its lexical complexity or even grammaticality. 

BabyLM and BabyBabelLM have introduced datasets that include training data that would be more likely for a child to encounter, such as child-directed speech, however participants have found that some of this data degrades performance on standard benchmarks \cite{padovani-etal-2025-dialogue}.

Optimizing a dataset through selection is a difficult, if not intractable, task. Instead, we propose a simpler method: use an existing pretrained Large Language Model (LLM) to improve the conciseness of the data. 
Pretrained LLMs have remarkable ability to summarize text, removing unimportant information such as filler words, or overly verbose descriptions. LLMs are also adept at translating between languages, allowing us to generate high-quality synthetic parallel training data. We experiment with both summarizing and translating the BabyLM Challenge's provided English data to Dutch and Chinese, the other two languages involved in the \textsc{Multilingual} track. 

\begin{figure}
    \centering
    \includegraphics[width=\linewidth]{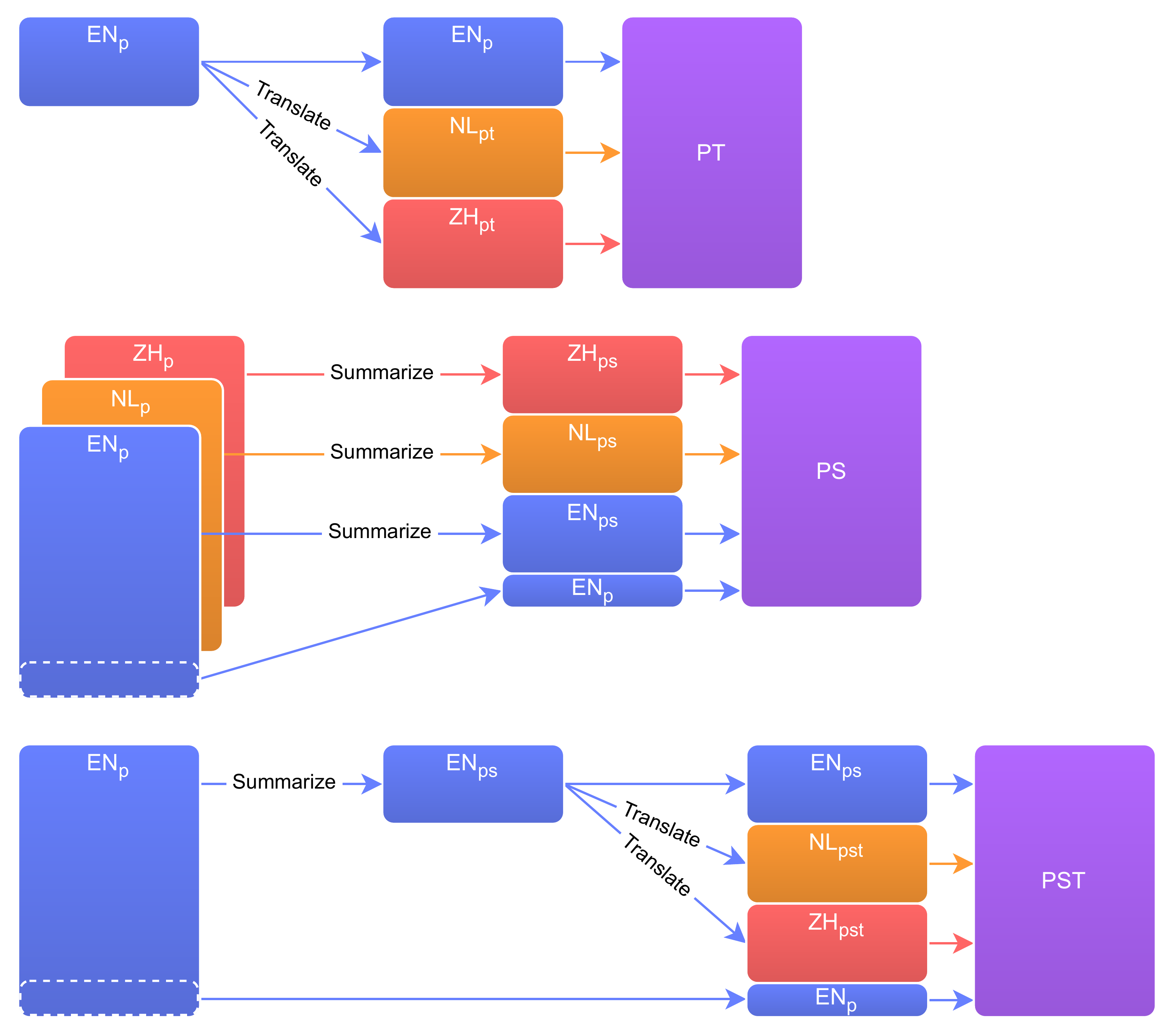}
    \caption{Preprocessing pipelines we investigate, including packed-translated (\textit{PT}), packed-summarized (\textit{PS}), and packed-summarized-translated (\textit{PST}). Packing process not shown for brevity.}
    \label{fig:pipeline}
\end{figure}

In this work, we find that:
\begin{itemize}
    \item Using 3 straightforward preprocessing steps, packing, summarization, and translation, provides a large performance improvement across all languages.
    \item Using a larger model is beneficial, providing more capacity for each language, but only when using LoRA for fine-tuning to prevent overfitting.
    \item Using a character-level tokenization also improves results in larger models, when using LoRA.
\end{itemize}

These three improvements culminate in an 11-point increase over the baseline, marking a substantially higher quality multilingual BabyLM.

\begin{table*}[]
    \centering \scriptsize
    \begin{tabular}{p{10cm}p{5cm}} \toprule
        Text & Summary\\ \midrule
        Well it's just that, you know, a pound, or a hundred pounds today, is not the same as a hundred pounds in a year's time, or two, two years' time. Right. So that would be opportunity cost? That's right, yes, that's right, so, in actual fact, this area here, right, is going to be the discounted sum of all future rural incomes. &  A pound today isn’t the same as a pound next year. This represents opportunity cost. The discounted sum of future rural incomes is represented here. \\ \midrule
        A: you can smell ether in the air sometimes  \newline
A: It keeps the neighborhoods happy I guess. \newline
B: I guess so. \newline
A: Uh, cause they clean the, uh, the lenses for cameras \newline
A: and they, they make film here \newline
A: but it's, you know, it's under so much control. \newline
A: You know, sulphur dioxide is the big emittant from them \newline
A: but that's really getting under control now. \newline
B: Right. \newline
B: I, you know, I don't know in air what they do. \newline
B: Uh, I, I haven't run across any major pollutants down there that I've, I've really seen a lot of, you know, about. \newline
B: Uh, I, I know I've seen like, uh, \newline
B: my grandparents live in Corpus Christi, Texas \newline
B: and I know they've, there's a lot of refineries down there \newline
B: and that, that's some pretty potent stuff they put up in the air. \newline
B: I, but I don't know how, uh, you know, \newline
B: there's a difference in what you can smell and what you, uh, you know, what's bad. \newline
A: Be interesting to see when, as Mexico develops industrially whether, you know, without & 
A: You can smell ether in the air sometimes. It keeps the neighborhoods happy. They clean camera lenses and make film. But it's controlled well. Sulphur dioxide is a big emitter, but it's getting under control now. \newline
B: I guess so. I don't know much about air pollution. I've never seen many major pollutants around. My grandparents live near Corpus Christi with lots of refineries, which can be potent. But smelling something doesn't mean it's harmful. It would be interesting to see how Mexico develops industrially. \\
        \bottomrule
    \end{tabular}
    \caption{Examples of original text and their respective LLM-generated summaries.}
    \label{tab:examples}
\end{table*}

\section{Related Work} \label{sect:related}

\subsection{Multilingual Pretraining}
Multilingual pretraining is typically approached with a similar strategy as monolingual pretraining: training an LM with masked or causal language modeling, with the only major change being an increased vocabulary size to better accomodate the diverse character sets. This was adopted as the standard since the release of mBERT \cite{devlin-etal-2019-bert}, and has remained effectively the same to this day. One of the latest examples of a performant multilingual model includes Qwen 3 \cite{yang2025qwen3}, which uses this same pretraining strategy, only with some architectural optimizations and adding a reasoning fine-tuning step.

\subsection{Synthetic Training Data}
A number of works have investigated the use of synthetic pretraining data for training models, both in and outside the BabyLM context. 
Some BabyLM submissions have tried to generate synthetic data on-the-fly, with limited success \cite{ulm-etal-2025-contrastive,alkhamissi-etal-2024-dreaming}.
\citet{theodoropoulos-etal-2024-berttime} and \citet{poh-etal-2025-say} generated synthetic data (bedtime stories and child-directed speech, respectively), finding this data negatively impacted overall performance.
\citet{edman-etal-2024-babylms} used a synthetic dataset that includes sentences and contrastive pairs, and showed improved results on GLUE tasks with a similar structure such as MRPC. 

Outside of BabyLM, \citet{eldan2023tinystories} created TinyStories, a synthetic dataset mimicking stories for 3-4 year olds, and found that small language models trained on this could generate fluent outputs with a limited vocabulary. 
Synthetically-generated textbooks have also been successfully used to pretrain a large-scale LLM \cite{gunasekar2023textbooksneed}.
\citet{kang-etal-2025-demystifying} built upon this and showed that synthetic data alone does not show an improvement in training efficiency, however a mixture of real and synthetic data does. 



\subsection{Tokenization}
Character-level models have been extensively explored, and have been shown to be more sample-efficient in training, despite their higher training cost overall due to longer sequence lengths \cite{xue-etal-2022-byt5}. They have been shown to be more performant on tasks requiring character-level manipulation \cite{pagnoni-etal-2025-byte}, and similar-language translation \cite{edman-etal-2024-character}.

In the context of BabyLM, character-level models have been seldom explored. \citet{edman-bylinina-2023-much} attempted using a character-level tokenizer without much success, and \citet{edman-fraser-2025-mask} used character-level information to inform an n-hot encoding, which showed improved adjective nominalization ability.

\section{Method}

We break down our method into the three improvements that we make over the baseline: preprocessing, tokenization, and model size.

\subsection{Preprocessing}
Our preprocessing involves 3 steps: packing, summarizing, and translation. 
For the latter two steps, we use \texttt{Qwen2.5-3B-Instruct} \cite{qwen2024qwen25}\footnote{\url{https://hf.co/Qwen/Qwen2.5-3B-Instruct}}. We show the prompts used in Figure \ref{fig:prompt} in Appendix \ref{app:prompt}. 

\paragraph{Packing}
Packing is simply done by concatenating examples together until they form a desired context length. We use a context length of 512 words. This is primarily done because the structure of examples in the datasets provided by BabyLM does not preserve document boundaries, so for most examples, the surrounding examples are from a related context. 

\paragraph{Summarization}
Summarization is done with the purpose of improving the conciseness of the dataset. Some of the data, particularly that which is transcribed speech, has several filler words or phrases (e.g., 'uh', 'you know'), or repetitions that do not add semantic value. While some of these are easily detectable with simple rules, others are more difficult to replace without understanding the context. We therefore use an LLM to shorten the text. 

Following the packing step, we feed the packed examples into the model, asking it to shorten the text to 200 words or less. 
We note that in our prompt we specify to `shorten' the text, rather than use the word `summarize'. This is because for chunks of dialogue, a summary may imply a 3rd-person overview of the conversation. We wish to preserve the dialogue format. 

We show examples of the summarization in Table \ref{tab:examples}. The summaries show a complete removal of filler words, resulting in a significant reduction of the text length. The summaries are however not perfect. The dialogue contains less interaction between the persons, and it even mistakenly attributes the last sentence to person B rather than person A. Nevertheless, we do not need the summarized text to be a perfect reflection of the original text, we only need it to be high-quality training data. 

\paragraph{Translation}
Translation is done to create parallel data. Training on parallel data is typical strategy in MT and generally leads to better alignment, but finding high-quality human-translated parallel data is difficult, particularly trilingual data covering English, Dutch, and Chinese. Meanwhile, LLMs can produce high-quality translations. We therefore translate our English data to Dutch and Chinese using the packed examples or the packed-summarized examples. 

With these example triplets, we then concatenate each triplet together, using language tags to separate the languages. We show an example below:

\begin{quote}
\texttt{The sun shines.
<dutch> De zon schijnt.
<chinese> \begin{CJK*}{UTF8}{gbsn}阳光照耀。\end{CJK*}}
\end{quote}

We vary the order of the languages, using all 6 permutations equally often. The goal of this format is aligned with the auto-regressive nature of a decoder model: The first part teaches the model to predict the next token based only on its knowledge of the first language, while the subsequent parts let the model use its understanding of the previous languages to inform what should come next. This encourages the model to form an internal mapping between languages. In practice, each part of the triplet consists of several sentences (using our summarized context size of 200 English words), so the model is effectively trained on something more akin to paragraph-level translation.

\paragraph{Pipeline}
Given these 3 preprocessing steps, we assemble them into different pipelines to ablate their effect. The pipelines include packed-only (\textit{P}), packed-translated (\textit{PT}), packed-summarized (\textit{PS}), and packed-summarized-translated (\textit{PST}). We visualize the latter three in Figure \ref{fig:pipeline}. 

The datasets we use are provided by the organizers: \texttt{BabyLM-2026-Strict}\footnote{\url{https://hf.co/datasets/BabyLM-community/BabyLM-2026-Strict}} for English, \texttt{babylm-nld}\footnote{\url{https://hf.co/datasets/BabyLM-community/babylm-nld}} for Dutch, and \texttt{babylm-zho}\footnote{\url{https://hf.co/datasets/BabyLM-community/babylm-zho}} for Chinese. \textit{P} and \textit{PT} require only one third of the datasets. 

For \textit{PS} and \textit{PST}, the summarization process reduces the data to less than 33\% of the original size, so we fill the remaining budget (6.2\% and 32.3\%, respectively) with packed English data.

\subsection{Model Size}
A natural question is whether the default model size has enough capacity to adequately support all 3 languages, while simultaneously preventing overfitting. To test this, we increase the hidden dimension size from 768 to 1024, intermediate size from 3072 to 4096, and number of layers from 12 to 24. This results in a roughly 4$\times$ larger model. 

The main obstacle for training a larger model, aside from additional compute, is an increased risk of overfitting. For the pretraining phase, this is less of an issue, given that the 100M word budget allows for diverse enough text. Finetuning, on the other hand, does become problematic. To handle this, we use LoRA \cite{hu2022lora}, which enables us to more conservatively update model parameters for a specific task. We use a LoRA rank of 256, alpha 512, 5\% dropout, and learning rate 1e-4, applying LoRA to all linear layers except the gating layer. 

\subsection{Tokenization}
Choosing an optimal vocabulary size is a nontrivial task, particularly for multilingual models, where one must balance tokens across languages. Our preliminary experiments found that overall, a vocabulary size of 20k was most effective. However, upon closer inspection, character-level models perform better on zero-shot tasks, but substantially worse on finetuning tasks, bringing down their overall average. 

While out-of-the-box subword models outperform character models, 2 simple fixes reverse this finding. The first fix is identical to our fix for increasing model size: use LoRA. The second fix is similarly simple, increase the default context length used in finetuning from 128 to 512, accounting for the increased sequence length caused by character-level tokenization.

\subsection{Experimental Setup}
We use the Qwen3 model architecture \cite{yang2025qwen3} for our experiments. The primary reason for this is that Qwen3 uses RoPE \cite{su2021roformer}, which provides a useful bias towards local attention, eliminating the need for a scheduled context size increase as done in \citet{edman-bylinina-2023-much}. 

We note the full hyperparameter details in Table \ref{tab:hyperparams} in Appendix \ref{app:hyperparams}.
Our evaluation uses the \textsc{Multilingual} track's zero-shot and fine-tune tasks. 

\begin{table}[!t]\centering\small
\begin{tabular}{lrrrrr}\toprule
Preprocessing &$\mathit{\emptyset}$ &\textit{P} &\textit{PT} &\textit{PS} &\textit{PST} \\\midrule
\multicolumn{6}{l}{\textit{Zero-shot}} \\
eng &57.8 &55.4 &55.8 &57.2 &\textbf{59.0} \\
nld &\textbf{59.9} &58.2 &54.0 &55.6 &56.5 \\
zho &51.9 &52.2 &49.2 &52.5 &\textbf{52.8} \\
\multicolumn{6}{l}{\textit{Fine-tune}} \\
eng &29.1 &41.0 &43.8 &40.9 &\textbf{45.8} \\
nld &32.3 &39.0 &39.5 &37.4 &\textbf{40.2} \\
zho &20.3 &36.5 &40.5 &41.7 &\textbf{42.5} \\ \midrule
avg &39.5 &45.8 &46.3 &46.4 &\textbf{48.5} \\
\bottomrule
\end{tabular}
\caption{Effect of the 3 preprocessing steps: Packing (\textit{P}), Summarization (\textit{S}), and Translation (\textit{T})}\label{tab:prep_results}
\end{table}

\begin{table}[!t]\centering\small
\begin{tabular}{lcccc}\toprule
Model size &\multicolumn{2}{c}{Base} & \multicolumn{2}{c}{Large} \\ \cmidrule(lr){2-3}\cmidrule(lr){4-5}
LoRA? & No & Yes & No & Yes \\
\midrule
\multicolumn{5}{l}{\textit{Zero-shot}} \\
eng &59.0 &59.0 &\textbf{59.5} &\textbf{59.5} \\
nld &56.5 &56.5 &\textbf{57.8} &\textbf{57.8} \\
zho &52.8 &52.8 &\textbf{52.9} &\textbf{52.9} \\
\multicolumn{5}{l}{\textit{Fine-tune}} \\
eng &\textbf{45.8} &45.5 &21.7 &45.0 \\
nld &40.2 &40.2 &21.4 &\textbf{41.3} \\
zho &42.5 &42.7 &20.3 &\textbf{44.7} \\ \midrule
avg &48.5 &48.5 &36.2 &\textbf{49.3} \\
\bottomrule
\end{tabular}
\caption{Effect of model size and using LoRA for fine-tuning.}\label{tab:size_results}
\end{table}

\section{Results}

\subsection{Data Preprocessing}
We show the effect of our preprocessing steps in Table \ref{tab:prep_results}. The packing step shows the largest positive effect, which makes sense as some of the examples are exceedingly short on their own, such as someone simply saying ``Right.'' The additional context allows a model to learn longer-distance relationships between sentences, which is useful in particular for the downstream tasks requiring finetuning.

Beyond packing, both translation and summarization on their own provide a positive effect, but their combination yields the most improvement. We suspect that translation on its own may be not as effective as summarization then translation because the data being translated is of lower quality, and this is improved by the summarization step. Meanwhile, the summarization on its own, while improving data quality, does not benefit from data being parallel.

\subsection{Model Size}
The effect of model size can be seen in Table \ref{tab:size_results}. The larger model has a higher zero-shot performance across all languages, but suffers from overfitting when fully fine-tuned. LoRA resolves this issue, increasing the fine-tuned performance on Dutch and Chinese particularly. 

\subsection{Tokenization} \label{analysis}
We show the impact of vocabulary size in Table \ref{tab:progression}. By default, character models outperform a larger vocabulary on zero-shot tasks for English and Dutch, likely due to a better ability to generalize between the two Latin-scripted languages. However, on finetuning tasks, with no changes, we see a drop in performance. Using a larger context size of 512 tokens and LoRA, the performance recovers. Although the fine-tuning results are still slightly lower for English and Chinese, we see later in Table \ref{tab:progression} that with a larger model size it outperforms the larger vocabulary for English. As a character-level model has fewer parameters dedicated to the embeddings and prediction layer (roughly 6-7$\times$ fewer), it may be the case that the base character-level model is too small to fully accomodate a character-level representation. More work would be needed to confirm this.

A character-level vocabulary for Chinese is likely less impactful since Chinese characters contain much more information on average, so our models are still effectively operating on a subword level for Chinese. Chinese characters can be decomposed into Kangxi radicals, and this may be a better equivalent for the character-level granularity of English and Dutch. We leave this for future work.

\begin{table}[!t]\centering\small
\begin{tabular}{lcccc}\toprule
&\multicolumn{2}{c}{20k} & \multicolumn{2}{c}{Char} \\ \cmidrule(lr){2-3}\cmidrule(lr){4-5}
LoRA & No & Yes & No & Yes \\
\midrule
\multicolumn{5}{l}{\textit{Zero-shot}} \\
eng &59.0 &59.0 &\textbf{60.9} &\textbf{60.9} \\
nld &56.5 &56.5 &\textbf{58.6} &\textbf{58.6} \\
zho &\textbf{52.8} &\textbf{52.8} &52.7 &52.7 \\
\multicolumn{5}{l}{\textit{Fine-tune}} \\
eng &\textbf{45.8} &45.5 &37.0 &44.3 \\
nld &40.2 &40.2 &33.0 &\textbf{41.0} \\
zho &42.5 &\textbf{42.7} &39.8 &42.1 \\ \midrule
avg &48.5 &48.5 &45.5 &\textbf{48.8} \\
\bottomrule
\end{tabular}
\caption{Comparison of 20k vocabulary to character-level vocabulary, and the effect of LoRA.}\label{tab:token_results}
\end{table}

\begin{table}[!t]\centering\small
\begin{tabular}{lcccccc}\toprule
Model & \textit{O.B.} &$\mathit{\emptyset}$ &\textit{P} &\textit{PST} &\makecell{\textit{PST}\\Large} &\makecell{\textit{PST}\\Char\\Large} \\\midrule
\multicolumn{7}{l}{\textit{Zero-shot}} \\
eng &56.4 &57.8 &55.4 &59.0 &59.5 &\textbf{61.4} \\
nld &57.8 &\textbf{59.9} &58.2 &56.5 &57.8 &59.1 \\
zho &51.3 &51.9 &52.2 &52.8 &\textbf{52.9} &\textbf{52.9} \\
\multicolumn{7}{l}{\textit{Fine-tune}} \\
eng &28.3 &29.1 &41.1 &45.8 &45.0 &\textbf{47.5} \\
nld &24.8 &32.3 &39.1 &40.2 &41.3 &\textbf{41.4} \\
zho &27.2 &20.3 &36.5 &42.5 &\textbf{44.7} &44.4 \\ \midrule
avg &38.9 &39.5 &45.8 &48.5 &49.3 &\textbf{50.2} \\
\bottomrule
\end{tabular}
\caption{Comparison of the official baseline (\textit{O.B.}) compared to various steps added, and final submitted model (right).}\label{tab:progression}
\end{table}

\begin{table}[!htp]\centering\small
\begin{tabular}{lccc}\toprule
Model &\textit{O.B.} &\makecell{\textit{PST}\\Char} &\makecell{\textit{PST}\\Char\\Large}\\\midrule
\multicolumn{4}{l}{\textit{Zero-shot}} \\
eng &53.6 &57.1 &\textbf{57.4} \\
nld &55.0 &55.3 &\textbf{56.1} \\
zho &47.3 &49.3 &\textbf{51.0} \\ \midrule
\multicolumn{4}{l}{\textit{Fine-tune}} \\
eng &43.1 &50.4 &\textbf{53.3} \\
nld &39.8 &47.7 &\textbf{48.1} \\
zho &40.7 &47.5 &\textbf{49.6} \\\midrule
avg &45.9 &50.9 &\textbf{52.4} \\ 
\bottomrule
\end{tabular}
\caption{Comparison of the official baseline (\textit{O.B.}) to our two leaderboard submissions, including hidden tasks. \textit{PST Char} and \textit{PST Char Large} are equivalent to \textit{Qwenfant-0.1B} and \textit{Qwenfant-0.4B}, respectively.}\label{tab:leaderboard}
\end{table}

\subsection{Overview}
We show our progression from the baseline scores to our final model in Table \ref{tab:progression}. Here, we can see the impact of our methods as a whole. The majority of the performance improvement comes from preprocessing steps, roughly 9 points. Using a larger model and character-level tokenizer only improve the performance marginally. The overall improvement of our model over the organizers' baseline is around 11 points. 

Our final model submissions, with averages including the added hidden tasks, is shown in Table \ref{tab:leaderboard}. We opted to submit our character-level models, one larger 0.4B model for optimal performance, and one smaller 0.1B model, as this is a standard size and roughly equivalent to the official baseline size. We can see that our preprocessing and tokenization strategy bring 5 points of improvement alone, and increasing the model size brings another 1.5 points.

\section{Conclusion}
When pretraining multilingual models on limited data, we see that we can improve on out-of-the-box methods in a few different ways. The main improvement we see comes from our preprocessing pipeline, where a pack-summarize-translate strategy is particularly effective at both providing higher-quality data and providing parallel data. Models trained on this data perform consistently well, showing that synthetic data can be valuable, if not better, than natural data. 
We also found that character-level models and larger models that typically suffer from overfitting can be coupled with a controlled fine-tuning using LoRA, thereby preventing this overfitting. 

These methods provide a sample-efficient pretraining strategy that improves performance on English, Dutch, and Chinese. Future work should extend this to further languages, in order to see if these strategies scale to a massively multilingual language model. It also remains to be seen if these methods have diminishing returns in a less sample-efficient setting. 

\section*{Acknowledgments}
The authors gratefully acknowledge the scientific support and HPC resources provided by the Erlangen National High Performance Computing Center (NHR@FAU) of the Friedrich-Alexander-Universität Erlangen-Nürnberg (FAU) under the NHR project b279bb. NHR funding is provided by federal and Bavarian state authorities. NHR@FAU hardware is partially funded by the German Research Foundation (DFG) – 440719683.
This work was co-funded by the European Union (ERC, EPICAL, 101141712). Views and opinions expressed are however those of the author(s) only and do not necessarily reflect those of the European Union or the European Research Council. Neither the European Union nor the granting authority can be held responsible for them.


\bibliography{acl_min,custom}

\appendix

\section{Training Hyperparameters} \label{app:hyperparams}

We show the complete hyperparameters used in Table \ref{tab:hyperparams}. Notably, we use the Muon optimizer \cite{jordan2024muon} along with AdamW for the embedding parameters (as Muon only works for 2D parameters). We chose Muon as it appears to have a better scaling factor \cite{qiu2026hyperparameter}.

\begin{table}[!htp]\centering\small
\begin{tabular}{lrrr}\toprule
&Parameter &Value \\\midrule
\multirow{5}{*}{Model} &Architecture &Qwen3 \\
&Hidden Size &768 \\
&Intermediate Size &3072 \\
&Layers &12 \\
&Vocabulary Size &20000 \\ \midrule
\multirow{12}{*}{Training} &Context Size &512 \\
&Batch Size (in tokens) &32768 \\
&Epochs & 10 \\
&Grdient Clipping &1 \\
&Warmup Steps &1000 \\
&Scheduler & Cosine \\
&Optimizers & Muon, AdamW \\
&Learning Rates & 2e-2, 5e-3 \\
&Weight Decays & 0.1, 0.01 \\
&Epsilons &1e-7, 1e-8 \\
&Muon Momentum & 0.95 \\
&AdamW Betas &0.9, 0.95 \\

\bottomrule
\end{tabular}
\caption{Hyperparameters used (except where otherwise indicated).}\label{tab:hyperparams}
\end{table} 

\section{Prompts} \label{app:prompt}
Figure \ref{fig:prompt} shows the exact prompts used for summarization and translation.

\begin{figure}[t]
\centering
\begin{tcolorbox}[
    colback=gray!5,
    colframe=gray!50,
    width=0.95\linewidth,
    boxrule=0.5pt,
    arc=2mm
]
\small
\textbf{User:} Translate the following text to \{\texttt{tgt\_lang}\}. Don't output any other text: \\
\\
\{\texttt{text}\}

\medskip
\textbf{Assistant:}
\end{tcolorbox}
\begin{tcolorbox}[
    colback=gray!5,
    colframe=gray!50,
    width=0.95\linewidth,
    boxrule=0.5pt,
    arc=2mm
]
\small
\textbf{User:} Shorten the following text to \{\texttt{sum\_len}\} words or less. Don't output any other text: \\
\\
\{\texttt{text}\}

\medskip
\textbf{Assistant:}
\end{tcolorbox}
\caption{Prompts used for translation and summarization.}
\label{fig:prompt}
\end{figure}

\end{document}